\documentclass{article}

\usepackage[dblblindworkshop, final]{neurips_2025}

\workshoptitle{Learning from Time Series for Health}

\usepackage{graphicx}
\usepackage{enumitem}
\setlist{nolistsep}
\usepackage{multirow}
\usepackage{tabularx}

\usepackage[utf8]{inputenc} 
\usepackage[T1]{fontenc}    
\usepackage{hyperref}       
\usepackage{url}            
\usepackage{booktabs}       
\usepackage{amsfonts}       
\usepackage{nicefrac}       
\usepackage{microtype}      
\usepackage{xcolor}         

\title{Towards Characterizing Knowledge Distillation of PPG Heart Rate Estimation Models}

\author{%
  Kanav Arora \\
  University of Washington\\
  Seattle, WA \\
  \texttt{kanava@cs.washington.edu} \\
  \And
    Girish Narayanswamy \\
  University of Washington \\
  Seattle, WA \\
  \texttt{girishvn@uw.edu}
  \AND
  Shwetak Patel \\
  University of Washington \\
  Seattle, WA \\
  \texttt{shwetak@cs.washington.edu} \\
  \And
  Richard Li \\
  University of Washington \\
  Seattle, WA \\
  \texttt{lichard@cs.washington.edu} \\
}

\begin{document}

\maketitle

\begin{abstract}

Heart rate estimation from photoplethysmography~(PPG) signals generated by wearable devices such as smartwatches and fitness trackers has significant implications for the health and well-being of individuals.
Although prior work has demonstrated deep learning models with strong performance in the heart rate estimation task, in order to deploy these models on wearable devices, these models must also adhere to strict memory and latency constraints.
In this work, we explore and characterize how large pre-trained PPG models may be distilled to smaller models appropriate for real-time inference on the edge.
We evaluate four distillation strategies through comprehensive sweeps of teacher and student model capacities: (1)~\textit{hard distillation}, (2)~\textit{soft distillation}, (3)~\textit{decoupled knowledge distillation~(DKD)}, and (4)~\textit{feature distillation}.
We present a characterization of the resulting scaling laws describing the relationship between model size and performance.
This early investigation lays the groundwork for practical and predictable methods for building edge-deployable models for physiological sensing.
  
\end{abstract}

\section{Introduction}

Wearable devices such as smartwatches and fitness trackers have enabled the collection of in-situ datasets of sensor signals with the potential to support individuals in tracking and monitoring their health and well-being.
Amongst other signals, photoplethysmography (PPG), a method for optical estimation of blood volume pulse (BVP), has shown utility in allowing individuals to gauge their cardiovascular health~\cite{pillai2024papagei, shabaan2020survey}.
The growing ubiquity of wearable devices has led to the accumulation of large PPG datasets~\cite{meier2024wildppg, Park2025GalaxyPPG, ppg-dalia_495} and the subsequent training of large neural models useful in estimating cardiac function such as heart rate and heart rate variability~\cite{narayanswamy2024scaling, pillai2024papagei, saha2025pulse, xu2025lsm, zhang2025sensorlm}.
These developments represent significant progress towards end-user applications, such as providing real-time feedback in exercise contexts (e.g.,~heart rate response to exercise intensity) as well as passive screening of diseases (e.g.,~hypertension).


Despite the success of these large models across a variety of sensor data tasks, their significant computational requirements pose a barrier to adoption and limit their utility~\cite{lee2025foundation}.
While edge models such as those running on wearables better preserve privacy and better support real-time feedback, large sensor models may struggle to realize these gains.
More work is thus needed to develop and characterize methods for enabling large physiological sensing models to effectively scale to the edge.


Prior work has established the utility of knowledge distillation~\cite{gou2021knowledge, hinton2015distillingknowledgeneuralnetwork}, where efficient student models learn from larger, high capacity, pretrained teacher models.
For example, DistilBERT~\cite{sanh2019distilbert} has found success in optimizing language models for edge deployments while retaining strong performance.
More similar to wearable physiological sensing, prior work has found success in distilling audio~\cite{peplinski2020frill} and accelerometer models~\cite{tang2021selfhar} useful for human activity recognition.
However, while knowledge distillation has been established as a powerful tool in developing compute-efficient models, there has been little exploration into the characterization of these methods, making it difficult to predict the performance of a distilled model.
Only recently have scaling laws that govern the distillation of language models been established to predictably compute distilled language model performance~\cite{busbridge2025distillation}.

Building off these ideas, our work takes a first step towards establishing predictable distillation performance in the domain of physiological sensing. Specifically, for the task of PPG heart rate estimation, 
we evaluate four distillation strategies across different student and teacher model capacities and characterize the effect of these variables on distilled model size.
We further compare the interplay between model computational requirements (i.e.,~memory consumption and inference time) and distilled performance.
We confirm that distilled models improve upon models trained from scratch, find that decoupled knowledge distillation outperforms other evaluated strategies, demonstrate that the performance of distilled models follow a characterization exponential scaling curve, and observe that these scaling behaviors vary by model architecture.

\section{Methods}

\textbf{Distillation Experiment Setup.}
We characterize the distillation scaling behavior of physiological sensing models for PPG across a number of teacher and student model sizes. 
Specifically, we employ the variant of a 1D-ResNet backbone~\cite{he2016deep} used by Meier et al.~\cite{meier2024wildppg} to classify the instantaneous heart rate given a PPG signal window.
To vary model capacity, we sweep across the number of residual blocks (i.e.,~resulting in an approximately exponential sweep of model parameters) for student and teacher models, as illustrated in Table~\ref{tab:parameters}.
We further explore the following four distillation strategies:

\begin{itemize}[leftmargin=*]
    \item \textit{Hard Distillation:}~The teacher model's predictions (i.e.,~the final $argmax$ output) are used as labels for training the student model, helping it mimic the discrete decision boundaries of the teacher.
    
    \item \textit{Soft Distillation:}~The student model is trained on the output probability distributions of the teacher model, encoding richer information about inter-class relationships and uncertainty~\cite{hinton2015distillingknowledgeneuralnetwork}.
    
    \item \textit{Decoupled Knowledge Distillation~(DKD):}~The teacher model's outputs are separated into target class and non-target class distillation components in the student model's loss to introduce flexibility in weighting the significance of true label and incorrect label probabilities~\cite{zhao2022decoupledknowledgedistillation}.

    \item \textit{Feature Distillation:}~Moving beyond operating on model outputs, in \textit{feature distillation}, the student model is trained to match the learned feature maps of the teacher model, aligning their intermediate representation spaces~\cite{romero2015fitnetshintsdeepnets}.
\end{itemize}

Heart rate estimation performance is evaluated via Mean Absolute Error (MAE) in beats per minute (BPM).
The performance of all distilled student models are evaluated against a corresponding model of the same size trained from scratch.



\begin{table}[t]
\caption{Experimental variables for characterizing the process of distilling PPG models.}
\begin{tabular}{lll}
\hline \hline \noalign{\vskip 3pt}
\textbf{Name} & \textbf{Description} & \textbf{Values} \\ \noalign{\vskip 1pt} \hline \noalign{\vskip 1pt}
Strategy & \begin{tabular}[c]{@{}l@{}}Procedure \\for distillation\end{tabular} & \begin{tabular}[c]{@{}l@{}}Hard Distillation, Soft Distillation,\\Decoupled Knowledge Distillation, Feature Distillation\end{tabular} \\ \noalign{\vskip 1pt} \hline \noalign{\vskip 1pt}
Teacher size & \begin{tabular}[c]{@{}l@{}}\# of residual blocks\\(\# of parameters)\\ in teacher model\end{tabular} & \begin{tabular}[c]{@{}l@{}}2 (33,724), 3 (44,156), 4 (54,588), 5 (97,852),\\ 6 (139,196), 8 (221,884), 10 (534,460), 12 (863,676)\end{tabular} \\ \noalign{\vskip 1pt} \hline \noalign{\vskip 1pt}
Student size & \begin{tabular}[c]{@{}l@{}}\# of residual blocks\\(\# of parameters)\\ in student model\end{tabular} & \begin{tabular}[c]{@{}l@{}}1 (23,292), 2 (33,724), 3 (44,156), 4 (54,588),\\ 5 (97,852), 6 (139,196), 8 (221,884), 10 (534,460)\end{tabular} \\ \hline \hline
\end{tabular}
\label{tab:parameters}
\end{table}

\textbf{Training Procedure.}
All models were trained for 300 epochs at a learning rate of $5 * 10^{-4}$ using a cross-entropy loss.
Following the task formulation given by Meier et al.~\cite{meier2024wildppg}, all models are trained to predict the instantaneous heart rate via classification by making a decision between 180 classes corresponding to heart rate values between 30 to 210 BPM.


\textbf{Datasets.}
For all experiments, we leverage three free-living PPG datasets containing a total of 107 hours of PPG sensor signals: (1)~WildPPG~\cite{meier2024wildppg}, (2)~PPG-DaLiA~\cite{ppg-dalia_495}, (3)~GalaxyPPG~\cite{Park2025GalaxyPPG}.
Following prior work, we use only the green channel of the PPG sensor, resampled to 25~Hz and segmented into 8-second windows with 2-second strides~\cite{meier2024wildppg, ppg-dalia_495}.
Each dataset includes heart rate ground truth (i.e., in BPM) derived via an electrocardiogram~(ECG) signal.
We generate participant-independent train-test splits by taking data from 80\% of the participants for training, and data from 20\% of the participants for evaluation.
We conduct 2-fold cross validation across all experiments.

\section{Results}

\begin{figure}
\begin{tabularx}{\linewidth}{@{}XX@{}}
  \includegraphics[width=\linewidth]{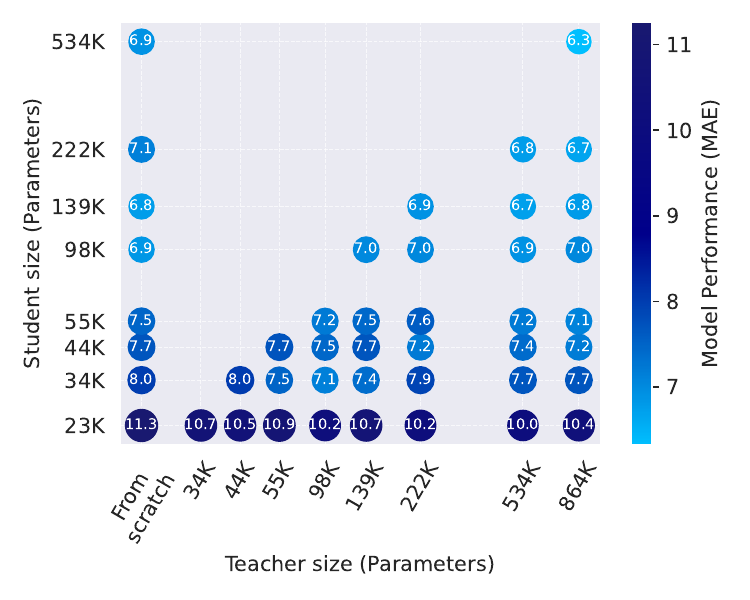}
  \caption{
    \textbf{\textit{DKD} Distilled Model Performance Across Different Student / Teacher Capacities.} 
    Color and size both encode MAE metric for instantaneous heart-rate prediction.
    The ``From scratch'' column denotes baseline models trained from scratch rather than  distilled from a teacher.}
    \label{fig:results1}
 &
  \includegraphics[width=\linewidth]{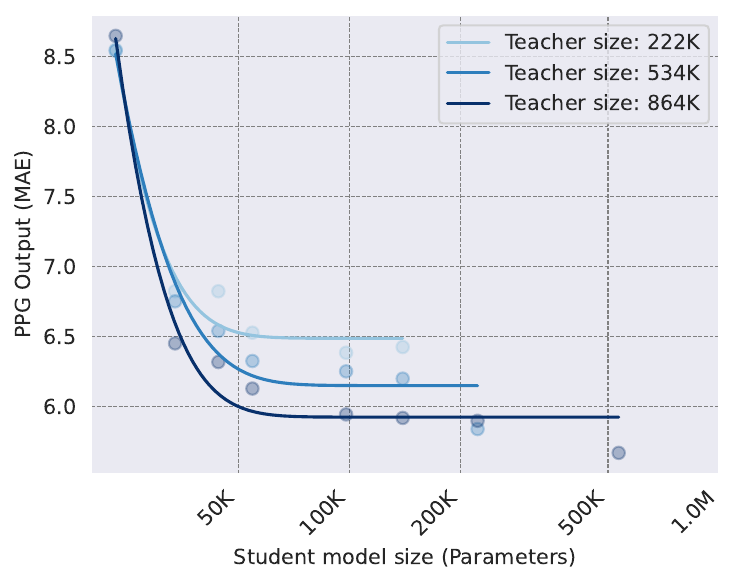}
  \caption{
    \textbf{\textit{DKD} Distilled Model Scaling Behavior.} 
    Scaling curves for distilled student model performance as a function of student and teacher model size.
    Note that experimental conditions with smaller teacher sizes yielded too few data points to effectively fit a curve.}
    \label{fig:results2}
\end{tabularx}
\end{figure}

\textbf{Distilled models outperform those trained from scratch.}
In Figure~\ref{fig:results1}, we show the results of our distillation experiment using the \textit{DKD} strategy.
The left-most column, ``From scratch'', denotes baseline models of a given size trained from scratch rather than distillation.
We find that this baseline is consistent with prior work (i.e.,~the target model size with 8-blocks yields a similar MAE to the results reported by Meier et al.~\cite{meier2024wildppg}  using the same model).
In general, we observe that smaller models exhibit worse MAE performance, and that distillation always improves performance over training from scratch.
We note that larger teacher models generally exhibit improved performance, and hypothesize that too-large models may overfit easily, resulting in degradation.
These results support that gains can be obtained in terms of performance from model distillation.

\textbf{\textit{DKD} outperforms other strategies.}
As shown in Table~\ref{tab:strategy_results}, we find that \textit{DKD} performs the best of the four strategies evaluated across all model size configurations, including across different teacher sizes not shown in the table.
\textit{DKD} is followed in performance by \textit{feature distillation}, then \textit{soft distillation}, and finally \textit{hard distillation}.
Out of the logit-based strategies, \textit{hard distillation} performed the worst due to the lack of information encoded in its discrete labels, and \textit{soft distillation} performed marginally better.
\textit{DKD}, on the other hand, poses the clearest advantage in being able to flexibly weigh true label and incorrect label probabilities, particularly in our task framing where the classification bins are semantically ordinal.
Through hyperparameter search, we found $\alpha=1$, $\beta=8$, temperature $\tau=2$, cross-entropy loss weight $CE=1$ to work best.
With these parameters, non-target class distillation~(NCKD) probabilities are weighed 8 times more than target class distillation~(TCKD) probabilities.
Our results indicate that while our small models may not have the capacity to learn a rich representation when trained from scratch.
Instead, by regressing to richer probability labels (via distillation) rather than the original BPM ground truth labels, the student models are able to more closely mimic and learn the internal representation of the larger teacher models, leading to improved performance.
We thus show that our student models are small enough to learn a strong internal representation independently given a rich enough label.

\begin{table}[t!]
\centering
\caption{Model performance (MAE) across different distillation strategies and student model sizes. Teacher model size is fixed at 12 blocks.}
\begin{tabular}{lcccccccc}
\hline
\hline
\noalign{\vskip 3pt}
\multirow{2}{*}{\textbf{\begin{tabular}[c]{@{}l@{}}Distillation\\ Strategy\end{tabular}}} & \multicolumn{8}{c}{\textbf{Student Model Size (Blocks)}} \\
 & \textbf{1} & \textbf{2} & \textbf{3} & \textbf{4} & \textbf{5} & \textbf{6} & \textbf{8} & \textbf{10} \\ \noalign{\vskip 1pt}\hline\noalign{\vskip 3pt}
\textbf{Hard} & 11.734 & 10.418 & 9.256 & 7.478 & 7.208 & 6.983 &6.830&6.493\\
\textbf{Soft} & 10.380 & 7.703 & 7.200 & 7.111 & 7.042 & 6.801 & 6.679 & 6.327 \\
\textbf{DKD} & \textbf{8.899} & \textbf{6.772} & \textbf{6.689} & \textbf{6.849} & \textbf{6.522} & \textbf{6.291} & \textbf{5.959} & \textbf{5.759} \\
\textbf{Feature} & 9.397 & 7.200& 6.952 & 6.914& 6.872 & 6.800 & 6.659 & 6.409 \\ \hline \hline
\end{tabular}
\label{tab:strategy_results}
\end{table}

\textbf{Varying model size exhibits predictable scaling.}
In Figure~\ref{fig:results2}, we show results from a preliminary experiment regarding characterizing distillation in the physiological sensing domain.
We first note that, consistent with prior work on the scaling laws for language model distillation~\cite{busbridge2025distillation}, these trend lines follow a predictable exponential curve in mapping the size of student models to their distilled performance.
We observe that performance seemingly begins to saturate at student models of size 6 residual blocks (139K parameters).
We also note that although this figure shows fit curves for results obtained using the \textit{soft distillation} strategy, the \textit{DKD} and \textit{feature distillation} strategies also adhered to these curves while the \textit{hard distillation} strategy produced a much sharper saturation at an earlier point (i.e.,~at a smaller model size).

\begin{figure}[h]
    \centering
    \includegraphics[width=0.75\linewidth]{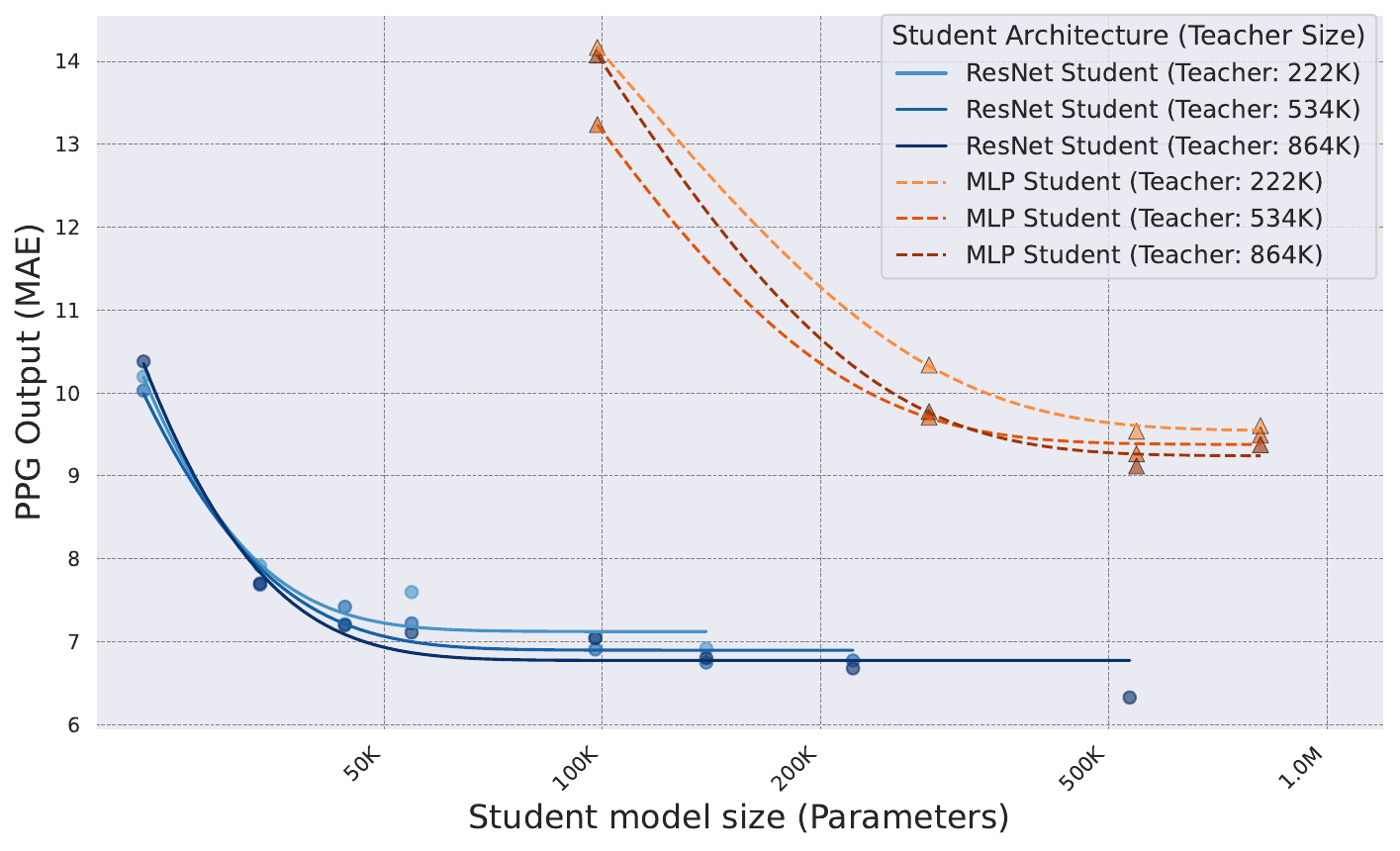}
    \caption{\textbf{\textit{Soft} Distilled Model Scaling Behavior for ResNet and MLP Student Architectures.} Performance analysis of student models trained via soft distillation across varying parameter counts. ResNet students (blue) demonstrate superior scaling efficiency and a significantly lower error floor compared to MLP students (orange), indicating a stronger inductive bias for the PPG task.}
    \label{fig:results_resnet_mlp}
\end{figure}

\textbf{Modeling decisions affect distillation scaling.}
We conduct an experiment to show the extent to which targeted modeling decisions affect distilled scaling by comparing the performance of distillation across different model architectures.
Specifically, we compare the scaling behavior of distilled student models with ResNet and multilayer perceptron (MLP) backbones, while consistently distilling from a ResNet-like teacher model.
Figure~\ref{fig:results_resnet_mlp} illustrates how MLP-based models, similar to ResNet-based models, also exhibit predictable scaling as a function of both teacher and student capacities.
However, we find that MLP-based student models consistently underperformed more sophisticated ResNet-based student models.
We hypothesize that the innate inductive biases of the convolutional layers (e.g., a natural tendency to smoothly filter signals) paired with more targeted architecture designs such as residual connections enable more sample-efficient learning.
We thus infer that while scaling may be observed for diverse model architecture, the specific scaling behavior may vary considerably.

\begin{table}[ht!]
\centering
\caption{
System compute benchmarks for distilled model inference.
Inference time is reported in seconds (mean $\pm$ standard deviation) and peak GPU memory is reported in megabytes.
}
\small 
\setlength{\tabcolsep}{3pt} 
\begin{tabular}{lccccccccc}
\hline\hline\noalign{\vskip 3pt}
\multirow{2}{*}{\begin{tabular}[c]{@{}l@{}}\textbf{Time \& Memory}\\ \textbf{Metrics}\end{tabular}} & \multicolumn{8}{c}{\textbf{Model Size (Blocks)}} \\
 & \textbf{1} & \textbf{2} & \textbf{3} & \textbf{4} & \textbf{5} & \textbf{6} & \textbf{8} & \textbf{10} & \textbf{12}\\ \noalign{\vskip 1pt}\hline\noalign{\vskip 2pt}
\textbf{Inference Time} (s)& 0.512 & 0.938 & 1.340 & 1.787 & 2.177 & 2.622 & 3.357 & 4.419 & 4.758 \\
& $\pm$0.025 & $\pm$0.028 & $\pm$0.0316 & $\pm$0.144 & $\pm$0.192 & $\pm$0.167 & $\pm$0.147 & $\pm$0.115& $\pm$0.130 \\
\noalign{\vskip 1pt}
\textbf{Memory Usage} (MB) & 9.468 & 9.646 & 9.824 & 10.002 & 10.623 & 11.275 & 12.568 & 18.440 & 23.483 \\
\noalign{\vskip 1pt}\hline \hline
\end{tabular}
\label{tab:strategy_results_squeezed_plain}
\end{table}

\textbf{Distillation can lead to large gains in memory consumption and inference time.}
Table~\ref{tab:strategy_results_squeezed_plain} shows system benchmarking of these models on an Nvidia RTX~2080-Ti GPU.
Although this is not representative of our final application scenario (e.g.,~microprocessors in wearable devices), we include these results to show the relative improvement made possible by distillation.
For example, distilling a 12-block model (i.e.,~the largest model we considered) to a 1-block model results in a nearly 90\% decrease in inference time and 60\% decrease in memory usage with only a 30\% reduction in MAE performance.

\section{Discussion and Conclusion}

\textbf{Dataset generalization.}
We presented an initial investigation into the distillation of heart rate estimation models.
Our evaluation used a naive cross-validation scheme with shuffling samples from three datasets.
We are interested in building off of work such as that of Kasnesis et al.~\cite{kasnesis2025replacing} towards further studying the generalizability of these distilled models across datasets (i.e.,~by training on one dataset and testing on another).

\textbf{Model architecture.}
Our preliminary investigation utilized a straightforward ResNet backbone model trained with supervision as the teacher model.
We are interested in continuing our experiments using larger models trained with more recent contrastive approaches (e.g.,~we note that the model in \cite{saha2025pulse} will be open source soon) to investigate how the potentially richer features learned in a self-supervised fashion might be distilled into smaller models.

\textbf{Novel distillation strategies.}
In this work, we leveraged four approaches to distillation already documented in the literature to provide baseline characterizations of these heart rate estimation models.
Leveraging insights from these experiments, we look forward to developing new methods of distillation that are particularly well-suited for this class of tasks.

This paper provides an initial demonstration of how knowledge distillation can be used to adapt large heart rate estimation models for resource-constrained wearable devices.
Our preliminary evaluation shows that distilled models consistently outperform those trained from scratch, with \textit{DKD} outperforming all other evaluated strategies.
We also characterized a scaling law that confirms distillation enables substantial reductions in memory usage and inference time for a modest trade-off in performance.
These findings provide an encouraging path forward for deploying powerful, real-time health monitoring models on the edge.

{
\small

\bibliographystyle{plain}
\bibliography{ref}

}


\end{document}